\documentclass[11pt,a4paper]{article}
\usepackage[hyperref]{acl2021}
\usepackage{times}
\usepackage{graphicx}
\usepackage{latexsym}
\usepackage{dialogue}
\usepackage{capt-of}

\usepackage{multirow} 
\usepackage{microtype}

\aclfinalcopy 

\title{Multi-Sentence Knowledge Selection in Open-Domain Dialogue}

\author{
Mihail Eric, Nicole Chartier, Behnam Hedayatnia, Karthik Gopalakrishnan,\\ 
{\bf Pankaj Rajan, Yang Liu and Dilek Hakkani-Tur}\\
\{behnam,karthgop,yangliud,hakkanit\}@amazon.com\\
Amazon Alexa AI
}

\date{}

\begin{document}
\maketitle
\begin{abstract}
Incorporating external knowledge sources effectively in conversations is a longstanding problem in open-domain dialogue research. The existing literature on open-domain knowledge selection is limited and makes certain brittle assumptions on knowledge sources to simplify the overall task~\cite{Dinan2019WizardOW}, such as the existence of a single relevant knowledge sentence per context. In this work, we evaluate the existing state of open-domain conversation knowledge selection, showing where the existing methodologies regarding data and evaluation are flawed. We then improve on them by proposing a new framework for collecting relevant knowledge,  and create an augmented dataset based on the Wizard of Wikipedia (WOW) corpus, which we call WOW++. WOW++ averages 8 relevant knowledge sentences per dialogue context, embracing the inherent ambiguity of open-domain dialogue knowledge selection. We then benchmark various knowledge ranking algorithms on this augmented dataset with both intrinsic evaluation and extrinsic measures of response quality, showing that neural rerankers that use WOW++ can outperform rankers trained on standard datasets. Our WOW++ dataset is open-sourced to further research in the area of knowledge selection.~\footnote{https://github.com/alexa/wow-plus-plus}
\end{abstract}

\section{Introduction}

One of the key components needed to enable robust and engaging open-domain conversational systems is the ability to select and integrate relevant knowledge from diverse knowledge sources. This knowledge selection setting is complex 
for a number of reasons: 1) relevant knowledge is highly dependent on conversational context, and requires understanding dialogue history and evolving user requirements at a turn-by-turn granularity, 2) because conversations are not topic-constrained, systems may need to pool knowledge from a theoretically boundless number of sources (i.e. the entire internet), and 3) knowledge may be represented in structured, semi-structured, and unstructured formats making it difficult to extract all the information needed for a conversation. 

\begin{figure}[t!]
    \centering
    \includegraphics[scale = 0.35]{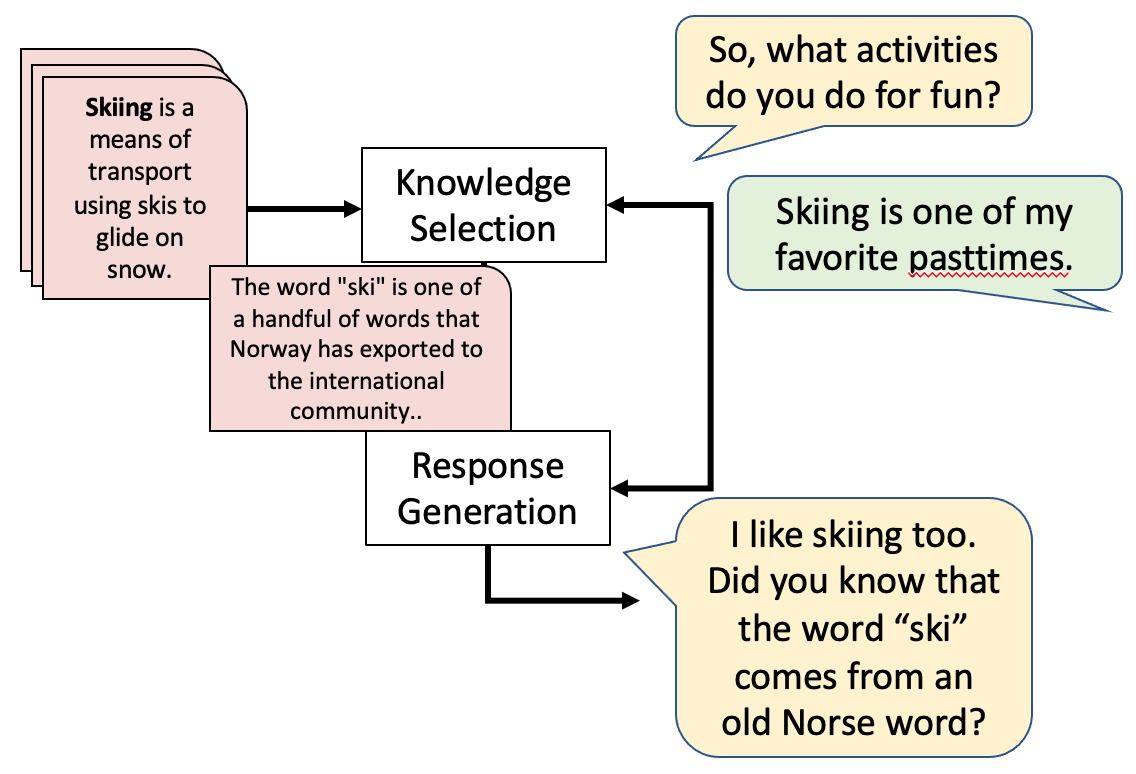}
    \caption{Overview of an end-to-end knowledge ranking and response generation system}
    \label{fig:architecture}
\end{figure}

Recently there has been increasing interest in the problem of knowledge selection for open-domain dialogue~\cite{ap3rd,Dinan2019WizardOW,Kim2020SequentialLK,Gopalakrishnan2019TopicalChatTK}. There have been numerous efforts to build standardized benchmark datasets whereby open-domain conversations are collected with crowdworkers who are given access to some closed set of knowledge sources such as a subset of Wikipedia or news articles~\cite{Dinan2019WizardOW,Gopalakrishnan2019TopicalChatTK}. These datasets suffer from a number of limitations including the fact that they either do not explicitly annotate relevant knowledge sentences per turn in the conversation or make the strict assumption that only a single utterance from a knowledge source is relevant. Without sufficient information about relevant knowledge for a given context, it is difficult to train data-driven models on the isolated problem of knowledge selection. 

In this work, we conduct a thorough analysis of the knowledge selection problem through the lens of the Wizard of Wikipedia (WOW) dataset, one of the standard knowledge-grounded open-domain dialogue benchmarks~\cite{Dinan2019WizardOW}. 
Our analysis qualitatively and quantitatively demonstrates that the strict one-knowledge-sentence-for-one-context assumption in the data is unreasonable, leading to a lack of meaningful interannotator agreement scores.

We then build on this result and relax this assumption, allowing for multiple knowledge snippets per context. We introduce a new continuously-scaled notion of relevance called \emph{wisdom-of-the-crowd-relevance} (WOC) and use this measure to reannotate about 800 dialog contexts from the original WOW corpus with relevant knowledge. This is done by taking a dialogue from WOW and extracting a subdialogue at a random turn in the dialogue. Our augmented WOW corpus, which we call WOW++, averages 8 knowledge sentences per dialogue turn, and demonstrates significantly more knowledge diversity.
Using WOW++, we then benchmark a number of different knowledge ranking algorithms using both standard information retrieval automatic measures as well as extrinsic human evaluation on generated responses. Our results indicate that neural rerankers using WOW++ are able to outperform other algorithms such as traditional IR baselines and neural models trained using the original WOW data.

\section{Related Work}

In recent years, knowledge selection in open-domain dialogue systems has seen a tremendous surge in interest as the community has recognized the utility of having these abilities in conversational systems~\cite{Ram2018ConversationalAT, Khatri2018AdvancingTS, ap3rd}. 

In one line of work, a number of industry research groups have demonstrated that large quantities of chat data coupled with the latest high-capacity Transformer-based models can produce particularly engaging and convincing conversational experiences ~\cite{Adiwardana2020TowardsAH,Roller2020RecipesFB}. While these models produce impressive outputs, they consciously shirk any explicit knowledge-selection mechanisms. Any knowledgeable appearance in their outputs tends to be a consequence of facts memorized in training data~\cite{lux2020truth}. In addition, the models have a tendency to generate facts that may be factually inaccurate, 
referred to as \emph{factual hallucination}. 

Knowledge selection in open-domain systems took a tremendous leap forward with the introduction of standard datasets of knowledge-grounded dialogues. The work of ~\cite{Zhou2018ADF,Dinan2019WizardOW,Gopalakrishnan2019TopicalChatTK} produced such corpora with upwards of 10K dialogues and up to 10s of dialogue turns leveraging knowledge from diverse sources such as Wikipedia, and the Washington Post. While certainly a step forward, these datasets introduced some unreasonable data constraints that aren't apt to the knowledge setting such as either no explicitly-annotated knowledge snippets or only a single one, making training of robust knowledge selection systems very difficult.

Since the introduction of these corpora, numerous groups have tackled the knowledge selection problem from different angles. For example, ~\citet{Kim2020SequentialLK} developed a sequential  latent  variable  model to help address ambiguity in knowledge sources in the WOW context. ~\citet{Zhao2020LowResourceKD} developed models that dealt with low-resource settings with general representations learned from ungrounded dialogues but finetuning done with small numbers of domain-specific training examples. ~\citet{Tuan2020UnsupervisedIO} recognized that even with external knowledge sources, there may be a knowledge gap that can be filled in real-time using unsupervised local knowledge. While these works created modeling improvements on existing datasets, there has still not been a study investigating how well-formed our existing datasets are.

\section{WOW++}

The WOW++ dataset we describe below is an augmented dataset based on the Wizard of Wikipedia (WOW) corpus ~\cite{Dinan2019WizardOW}. The WOW corpus consists of 22,311 dialogues containing 201,999 turns. The dialogues are comprised of two interlocutors who engage in chit chat on a given topic where one interlocutor is a knowledgeable expert in the topic. The expert, or wizard, is provided access to knowledge snippets from Wikipedia that are potentially relevant to a given conversation topic, asked to select one of the knowledge snippets and utilize the information from the knowledge snippet in their response. Thus, for a given wizard utterance only a single knowledge snippet is selected from a set of potentially relevant knowledge snippets. This selected snippet is considered to be the ground-truth, referred to as the \textit{gold} sentence, and is referred to as such throughout this paper.   

\subsection{Data Motivation: Limitations of One Knowledge}

\begin{figure}[t]
  \centering
    \includegraphics[scale = 0.65]{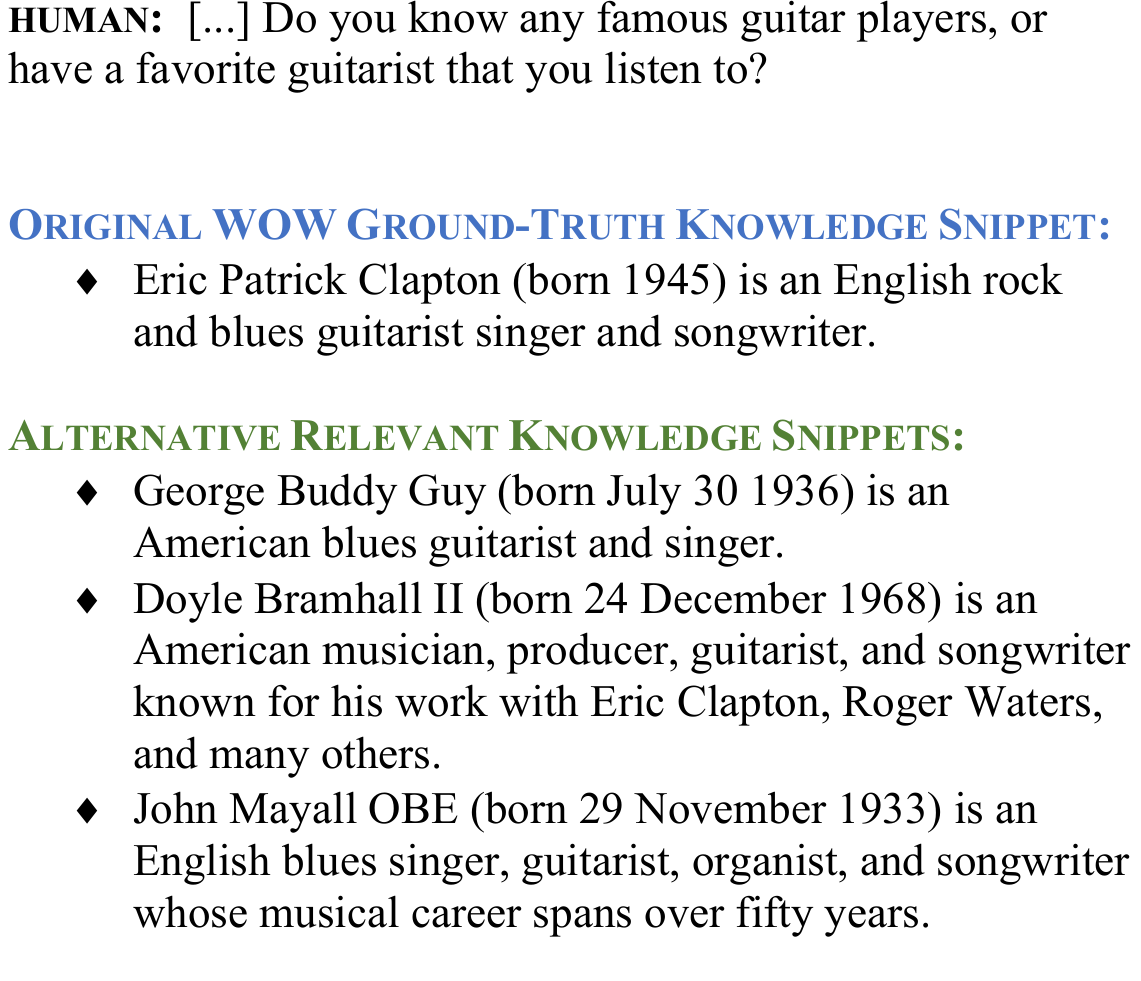}
    \vspace{-3mm}
    \caption{Example dialogue in WOW with ground-truth and alternative relevant knowledge snippets.}
    \label{fig:ericCLapton}
    \vspace{-3mm}
\end{figure}

In order to address the well-formedness of existing datasets, we identify two inter-related aspects of knowledge selection in open-domain dialogues:  the potential for multiple knowledge snippets to be relevant for a given response, and the subjective nature of relevance selection. Figure \ref{fig:ericCLapton} exemplifies how multiple knowledge snippets can be equally relevant for a specific question. All knowledge snippets, both the ground-truth snippet and the alternatives, come from Wikipedia. 
In this example, the response and relevant knowledge that could be leveraged in the response, is open to any guitarist. While the original WOW corpus identifies the knowledge snippet containing information about Eric Clapton as the ground-truth one (i.e. the \emph{gold} sentence)~\cite{Dinan2019WizardOW}, the alternative relevant knowledge snippets are equally relevant. There is nothing inherently more relevant with the Eric Clapton knowledge snippet than these alternatives. The choice, then, for one being \emph{the single relevant} knowledge snippet, would seem to be a reflection of personal preference of the annotator rather than objective relevance. 

\begin{figure}[t]
  \centering
    \includegraphics[scale = 0.65]{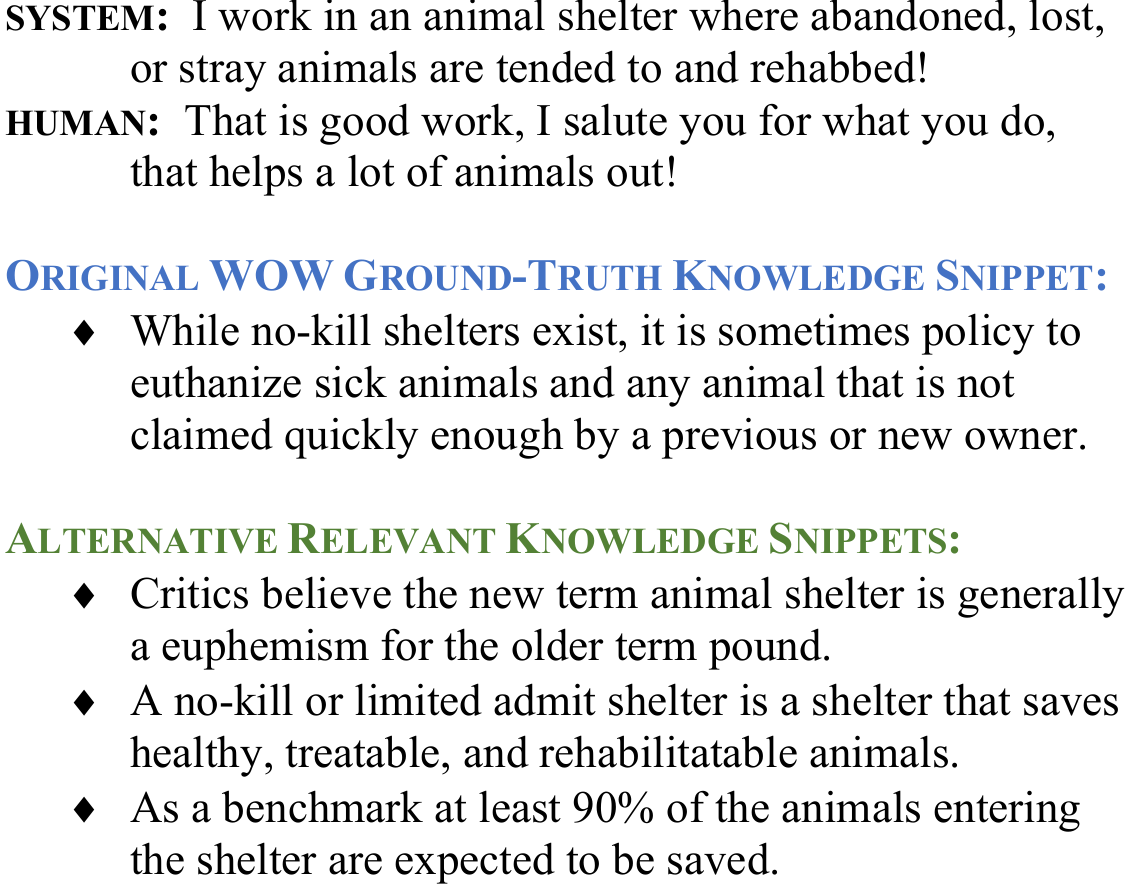}
    \vspace{-3mm}
    \caption{Example dialogue in WOW with ground-truth and alternative relevant knowledge snippets.}
    \label{fig:animalShelter}
\end{figure}

Annotator subjectivity is not only reflected in questions, but also in open-ended, general statements. Figure \ref{fig:animalShelter} depicts this scenario, in which the human's turn leaves the response by the assistant open:  there is no direct or leading question to limit the scope of the conversation. In this context, it would be just as reasonable for the system's next turn to leverage the ground-truth knowledge snippet provided by the WOW dataset as it would to leverage the alternative ones shown. 

In these cases, the choice of one knowledge sentence over all others reflects a subjective choice by the annotators, who are influenced by their own preconceived notions and expectations of the trajectory of the conversation. As such, although limiting relevance selection to a single knowledge sentence may be beneficial to simplify the annotation task, it does so at the expense of variation that is inherent and expected in open-domain dialogue. Further, the selection of one knowledge snippet in instances where many are relevant reflects annotator preferences, creating an issue of reproducibility. 

\subsection{Subjectivity of Knowledge Selection}

In order to address the feasibility of an annotation task in which multiple knowledge sentences are selected, we conduct a pilot study, adapting the methods used in the WOW study~\cite{Dinan2019WizardOW}.
We use a static evaluation that allows annotators to select multiple relevant knowledge snippets for only the final turn of a dialogue. This pilot study includes 40 in-house annotators selecting knowledge snippets for 20 dialogues from the WOW corpus. Annotators are provided with the dialogue and approximately 30 potentially relevant knowledge snippets, including the WOW-identified ground-truth knowledge snippet~\cite{Dinan2019WizardOW}, also from the WOW corpus. These 30 snippets come from, on average, 5 different Wikipedia articles. Annotators are instructed to select all knowledge snippets that could be used in the following turn to craft a relevant response, using the dialogue context to help assess relevance, and that a relevant knowledge sentence should not change the topic of the conversation. Figure~\ref{fig:ks_instructions} in Appendix shows the annotation instruction and interface. Please note that with this method, a single turn for each dialogue is annotated for knowledge selection.

We find that within the 610 snippets in the pilot annotation, 177 were not selected by a single annotator, i.e. not relevant and only 7 knowledge snippets were selected by all annotators as relevant. Further, we find that only 1 of the 20 gold sentences is selected by all annotators as relevant, the average proportion of annotators that select a gold sentence as relevant is 0.77. 

We first assess the quality of the pilot data by computing Fleiss' Kappa to measure inter-annotator agreement (IAA). We compute IAA for relevance selection in the pilot dataset (610 knowledge snippets, assessed by 40 annotators); agreement is moderate ($\kappa$ = 0.477, p $<$ 0.01)~\cite{landis1977measurement}. This finding is not surprising, as we outlined in the previous section that we suspected there would be subjective reactions to knowledge snippets. Next, we explore whether IAA would increase if only the WOW gold snippets are assessed for agreement. To do this, we create a subset of the pilot dataset, consisting of only the relevance assessments by our annotators of the WOW ground-truth knowledge snippets~\cite{Dinan2019WizardOW} and then compute Fleiss' $\kappa$ on the subset (20 knowledge snippets, assessed by 40 annotators). IAA for the ground-truth sentences is poor ($\kappa$ = 0.06, p $<$ 0.01)~\cite{landis1977measurement}. Finally, we examine whether low IAA was a reflection of some annotators' understanding of and ability to complete the task. 
We would expect that if the quality of some annotators' work were subpar, IAA should increase if we find a better subset of annotators. To assess this, we create 100 random samples of 10 annotators from the set of 40 and computed Fleiss' $\kappa$ for each sample. Agreement in these samples of 10 ranges from fair ($\kappa$ = 0.40, p $<$ 0.01) to moderate ($\kappa$ = 0.59, p $<$ 0.01). These results demonstrate that we are unable to create a subset of 10 annotators that agree on relevance selection. 

Although low IAA can be an indication of unreliable data~\cite{krippendorff1980validity}, it can also be an indication that the task is subjective~\cite{salminen2018inter}. We argue that low IAA in this context is a result of the inherent subjectivity of the knowledge selection task. Rather than clear and mutually exclusive categories, the notion of relevance in this context has fuzzy boundaries and can be dependent on the individual making the assessment. While there will be some agreement on relevance, it should not be assumed that relevance is an agreed-upon concept. Thus, rather than relying on absolute agreement among all annotators for relevance, we suggest that the notion of relevance be considered on a continuum. 

\subsection{Data Methodology:  Wisdom-of-the-Crowd Relevance}
Due to the limitations outlined above, we propose that knowledge selection be handled by appealing to the crowd. Using the same data collection approach as outlined in the pilot study, we conduct a knowledge selection task in which 798 dialogues and corresponding knowledge sentences were randomly sampled from the test seen (198), test unseen (200), and train (400) datasets in the original WOW corpus~\cite{Dinan2019WizardOW}. In order to make the task reasonable for annotators, we create 80 tasks on a crowd-source worker platform in which annotators are presented 10 randomized dialogues. As described above, for each dialogue a single turn is annotated for knowledge selection. 10 annotators assess knowledge snippets for these 798 dialogues, with an average of 30 knowledge snippets per dialogue. 
In order to mitigate low-quality annotations, we include two filter metrics in the task.   

Determining the threshold for relevance, or \emph{wisdom-of-the-crowd-relevance}, consists of a mixed approach of relevance vote distribution and manual inspection. The relevance vote for each knowledge sentence represents the proportion of annotators that selected a given knowledge sentence as relevant. We order the knowledge snippets from all dialogues by this relevance vote and use the third quartile of the vote distribution as the cutoff for relevance, resulting in a relevance threshold of 0.6. A sample of relevant knowledge sentences is manually inspected to ensure the quality and accuracy of the \emph{wisdom-of-the-crowd-relevance}. This approach to relevance accounts for variation due to inherent subjectivity while limiting noise expected from human evaluation.  

\begin{figure}[h!]
    \centering
    \includegraphics[scale = 0.21]{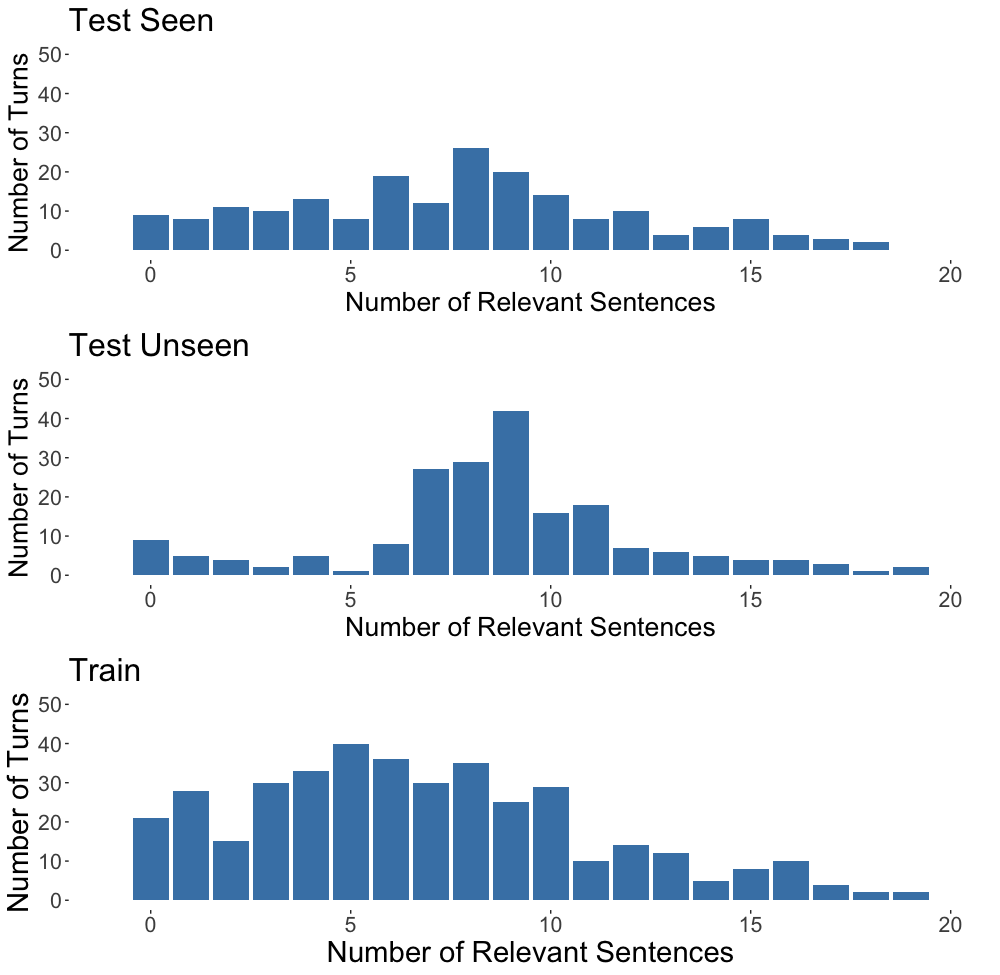}
    \caption{Histograms of the frequency counts for different numbers of relevant snippets per turn annotated from the test seen, test unseen, and train data.}
    \label{fig:DistrRel}
\end{figure}

The \emph{wisdom-of-the-crowd-relevance} scoring results in an average of 8 selected knowledge snippets per turn. Table \ref{tab:relevanceDistribution} provides a summary of relevant knowledge snippets in the WOW++ dataset. Figure \ref{fig:DistrRel} shows the distribution of relevant knowledge for the WOW++ dataset. Only 5\% of the dialogues in the dataset contain a single relevant knowledge snippet. These results suggest that, for the majority of dialogues, more than one knowledge sentence is relevant, and more importantly, that only a single knowledge sentence being relevant is the \emph{exception}, not the norm.  

\begin{table}[ht!]
    \centering
    \begin{tabular}{|c|c|}
    \hline
    Number of Dialogues & 798\\
    \hline
      Average Relevant KS per Turn &  8\\
      \hline
      Turns with no Relevant KS & 39\\
      \hline
      Turns with 1 Relevant KS & 41\\
      \hline
      Turns with $>$1 Relevant KS & 718 \\
      \hline
    \end{tabular}
    \caption{Counts of relevant knowledge snippets (KS) in WOW++}
    \label{tab:relevanceDistribution}
\end{table}

 To better understand the conversational contexts that have multiple relevant knowledge snippets, we examine a sample of the data to categorize the users' last utterance. Table \ref{tab:conversationCategories} depicts the three broad categories:  personal questions; factual questions; and general statements, and their distribution across our sample. Overall, the most frequent type is when the user's last utterance was a general statement. We then examined whether the multiple knowledge snippets came from the same topic (i.e. Wikipedia article) or were spread across the five different topics presented. Approximately 50\% of the relevant knowledge comes from a single topic for all three categories. For both general statements and factual questions, the remaining 50\% of knowledge is spread across 2 - 5 topics. For personal questions, 40\% of the knowledge comes from only two topics, and the final 10\% comes from 5 topics.

\begin{table}[h]
    \centering
    \begin{small}
    \begin{tabular}{c|c|c} \hline
        Category & Example & \%  \\
        \hline
         Personal  &  I love to play football, do you? & 10\\
         \hline
         Factual  & What else do you know about rap? & 33\\
         \hline
         General  & Soap operas seem so bad to me. & 57\\
         \hline
    \end{tabular}
    \end{small}
    \caption{Types of last user utterances and percent of occurrence in our dataset.}
    \label{tab:conversationCategories}
\end{table}

\begin{figure}[t]
    \centering
  \centering
    \includegraphics[scale = 0.65]{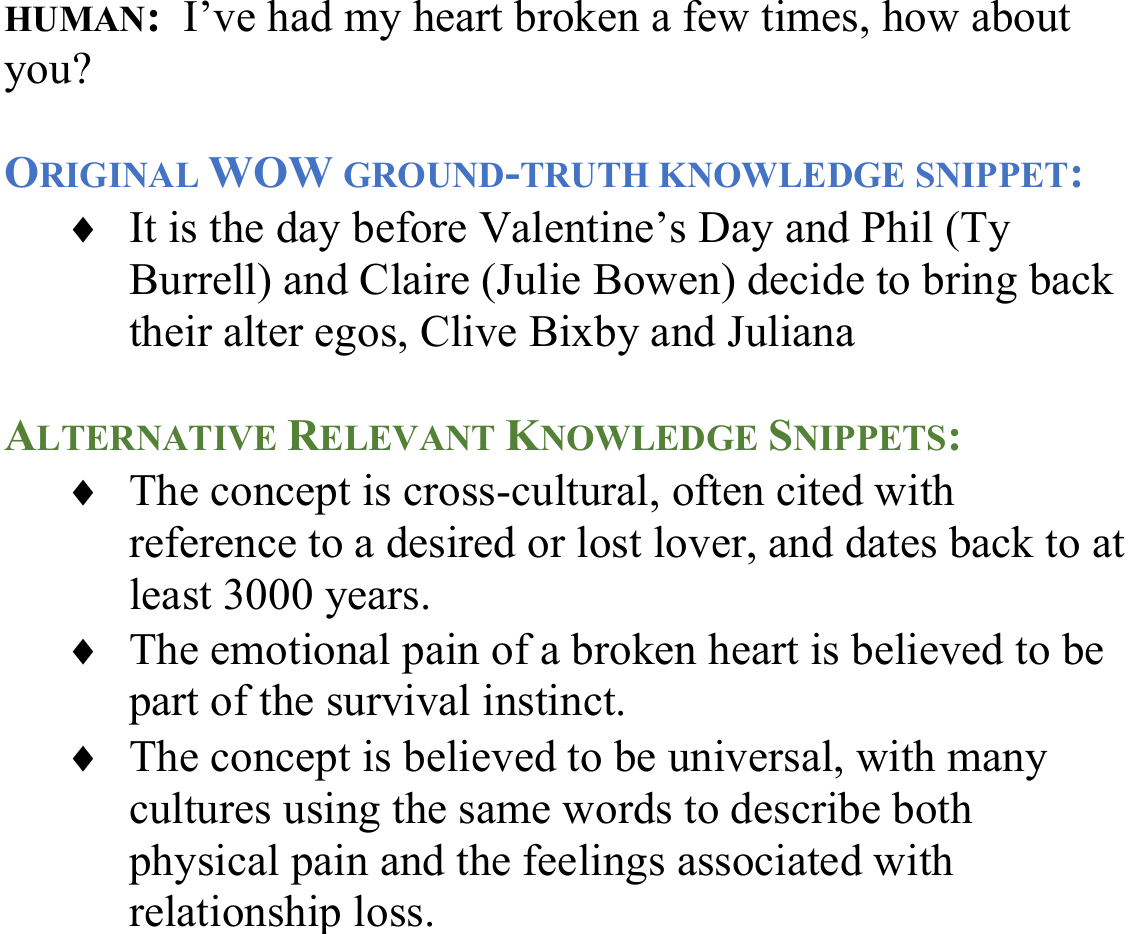}
    \caption{Example dialogue where WOW original ``gold'' is not relevant, but alternative snippets are.}
    \label{fig:badGold}
\end{figure}

Finally, we examined the original WOW gold snippet in relation to our multiple knowledge selection method. 
After removing dialogues where gold snippets were not included among the 30 candidates, there are a total of 697 conversations where the gold snippet is presented among the potentially relevant knowledge sources. Of those, there are 10 dialogues where the gold snippet is the only relevant knowledge snippet. However, there are 160 dialogues where the gold snippet does \emph{not} meet the relevance threshold. Although the gold snippet is not relevant in these 160 conversations, 136 have at least one alternative knowledge snippet selected as relevant. We inspect these instances and find that, in general, it is due to noise in the original WOW dataset. The dialogue in Figure \ref{fig:badGold} exemplifies this noise. Reading the WOW gold snippet, it is not clear how the knowledge in that snippet could be leveraged accurately to craft a relevant response to the question about heartbreak. This suggests that while a single person was able to craft a relevant response from this snippet, in general, it would not be seen as relevant. In other words, although the gold snippet is not relevant in these conversations, other knowledge snippets are, suggesting that there are knowledge snippets that are more ``gold'' than the WOW gold snippet.

By allowing for multiple relevant knowledge and introducing  \emph{wisdom-of-the-crowd-relevance} scoring, we have produced a robust augmented dataset that embraces the variation present in open-domain knowledge selection. We have demonstrated the assumption of one-knowledge-snippet-per-context needs to be re-assessed, as our data suggests that a single relevant knowledge snippet may not be reasonable nor replicable. Not only does this method help to mitigate some noise present in the original WOW dataset~\cite{Dinan2019WizardOW}, but we expect that it will be more fruitful when incorporating knowledge sources beyond Wikipedia.

\section{Method}
\label{sec:method}
We evaluate WOW++ using two different regimes to see the effect of the augmented knowledge snippets. The first is intrinsic evaluation for the knowledge reranking or selection task using automatic measures. The second is extrinsic evaluation where we provide a selected knowledge candidate to a response generation model and perform human annotation of system generated responses.  

\subsection{Knowledge Selection}
This intrinsic evaluation is intended to assess the performance of different models within the context of an isolated knowledge selection module. More formally, assume we are given a dialogue context consisting of $m$ turns represented as $D=\{T_1, T_2, ..., T_m\}$, and for the last turn, a list of knowledge sentence candidates are provided, $C=\{s_1, s_2, ..., s_n\}$, 
the system is asked to assess the relevance of these candidates and generate a ranked list, i.e.,   $f(C,D)=\{s_{i_1}, s_{i_2}, ..., s_{i_m}\}$.

We first evaluate using unsupervised sentence similarity scores for knowledge sentence relevance ranking, with scores calculated using traditional Tf-idf method or large pretrained models. 

\vspace{-0.05in}

\begin{itemize}
    \item \textbf{Tf-IDF}: Here we separately encode the dialogue context and each knowledge sentence from the set $C$ using term-frequency (TF) inverse-document-frequency (IDF) weights and then return the top 10 sentences that are most similar to the context vector using the cosine similarity metric. The IDF weights are learned from the full train set of the new WOW++ dataset by treating each sentence of the knowledge passages and dialogue turns as documents.
   
  \vspace{-0.05in} 
    \item \textbf{Static RoBERTa Retriever}: Here we encode the context and knowledge sentences using an off-the-shelf non-finetuned RoBERTa base model~\cite{Liu2019RoBERTaAR}. We then compute cosine similarity and return the top-ranked sentences.
    
\vspace{-0.05in}
    \item \textbf{MLM tuned RoBERTa Retriever}: Here we use the same scheme as in the \emph{Static RoBERTa} case except that the RoBERTa weights are first finetuned using a masked language modeling (MLM) objective on the WOW train set. This is analogous to the self-supervised pretraining described in ~\cite{Mehri2020DialoGLUEAN}.

\end{itemize}

\begin{table*}[t]
\centering
\resizebox{\textwidth}{!}{
\begin{tabular}{|l|l|l|l|l|l|l|l|}
\hline
\multicolumn{2}{|c|}{} & \textbf{MRR@1} & \textbf{MRR@5} & \textbf{MAP@5} & \textbf{MAP@10} & \textbf{NDCG@5} & \textbf{NDCG@10} \\ \hline
\multicolumn{2}{|l|}{TF-IDF}         &    0.66/0.74	& 0.76/0.84	& 0.56/0.65	& 0.57/0.63 &	0.80/0.87 &	0.81/0.86 \\ \hline
\multirow{2}{*}{Roberta Retriever} & Static RoBERTa & 0.51/0.58 & 	0.66/0.68 &	0.39/0.45 &	0.36/0.38	& 0.70/0.73	& 0.72/0.75 \\ \cline{2-8}
& Finetuned RoBERTa & 0.68/0.71 &	0.78/0.82 &	0.55/0.64	& 0.53/0.59	& 0.81/0.85	& 0.81/0.85 \\ \hline
\multirow{4}{*}{Roberta classifier}& WOW ++ & \textbf{0.75/0.84} &  \textbf{0.83/0.89}  & \textbf{0.67/0.75} & \textbf{0.67/0.73} & \textbf{0.86/0.90} & \textbf{0.86/0.90} \\ \cline{2-8}
 & WOW-original & 0.65/0.71 & 0.76/0.81 & 0.56/0.63 & 0.55/0.60 & 0.79/0.85 & 0.81/0.86 \\ \cline{2-8}
& WOW-original-subset & 0.30/0.45	& 0.46/0.59 & 	0.25/0.37 &	0.27/0.37 &	0.52/0.65	& 0.58/0.68 \\ \cline{2-8}
& WOW-original + WOW++ & 0.71/0.82 &	0.80/0.88 &	0.62/0.70	& 0.63/0.67 &	0.82/0.89	& 0.83/0.89 \\ \hline
\end{tabular}}
\caption{Knowledge selection: automatic evaluation results by data split (Test Unseen/Test Seen).}
\label{tab:automatic-eval}
\end{table*}

\vspace{-0.05in}
In addition, we propose to use supervised learning to train knowledge rerankers based on the pretrained language models, RoBERTa.   
We fine tune the RoBERTa base model by scoring an input sequence consisting of a dialogue context and a candidate knowledge sentence using a binary log-loss objective. 
This trained RoBERTa reranker outputs a binary probability of relevance for the context and candidate pair, which is used to rank the candidates.
In the following, we describe different training configurations in order to examine their effect on knowledge selection performance. 
For all the RoBERTa models, we use a maximum length of 256 tokens for the concatenated dialog context and knowledge candidate. 

\vspace{-0.05in}
\begin{itemize}
    \item \textbf{Training on WOW++}: we finetune a RoBERTa base model on WOW++ data. In training, a candidate knowledge sentence is considered positive if it exceeds our WOC relevance threshold.

    
    \vspace{-0.05in}
 \item \textbf{Training on WOW-original}:  the original WOW train set is used as the labelled data. Here we leverage one positive candidate (the gold knowledge snippet) and one negative candidate (sampled from the remaining knowledge) per dialogue context. Note that this scheme has about 120K examples, roughly an order of magnitude more data for training compared to only using WOW++. 

\vspace{-0.05in}
\item \textbf{Training on WOW-original-subset}: here we use the same dialogs as used in WOW++,
however, rather than using the new annotated snippets from WOW++, we use the gold knowledge snippet from the original WOW dataset corresponding to each context of the 400 training dialogues. Since this only introduces a single positive snippet per context, we additionally sample enough negative candidates from the available knowledge in each context so that the total samples used for training match the number used in the case when training with WOW++ data. 
    
    \vspace{-0.05in}
\item \textbf{Training on WOW-original + WOW++}: Here the training data contains both the WOW++ and the original WOW train set.

\end{itemize}

\subsection{Response Generation}
The extrinsic evaluation is intended to evaluate the knowledge selection module in the context of a fully-integrated dialog response generation system, thereby giving us a better understanding of end-to-end performance. We seek to understand the effect of different quality knowledge sentences on the downstream task of response generation.

Here, we first finetune the language modeling head of a GPT2 medium-model~\cite{Radford2019LanguageMA} on the original WOW corpus, using a similar setup as ~\cite{Wolf2019TransferTransfoAT} where the ground-truth knowledge and response are used for teacher-forcing training. The context is truncated to 64 tokens.
During inference, we take the top ranked sentence from a knowledge reranking model, use it along with the dialog context as a concatenated input to generate a response. 

\section{Experimental Results}



\subsection{Knowledge Selection Results}
For knowledge selection, we evaluate our models using a number of standard information retrieval metrics: MAP (mean average precision) and NDCG (normalized discounted cumulative gain) for the 5 and 10 candidate decision thresholds, and MRR (mean reciprocal rank) for the 1 and 5 candidate decision thresholds. 

Table \ref{tab:automatic-eval} shows the results for the knowledge selection methods presented in Section~\ref{sec:method}. 
First for the similarity-based methods, we see the traditional TF-IDF measure has strong performance. This may also speak to the manner in which the WOW data collection was done whereby crowdworkers could have optimized for snippets with high degrees of overlap with the dialogue context rather than necessarily those with the highest levels of semantic relevance. This is certainly an artifact of the original WOW data collection process where candidate articles were chosen via their TF-IDF weighted n-gram scores with the dialogue context.
Using the static RoBERTa model to computer similarity does not perform as well as the TF-IDF metrics, again, partly because of the reason discussed above.
Adapting the RoBERTa model to the task domain using the WOW data in an unsupervised fashion via MLM loss improves the performance significantly over the static models.
This is not surprising since the model is further trained on the domain data, resulting in better representations for words and sequences for similarity measures.



Regarding the RoBERTa classifier ranking approaches, results show that the model trained using the WOW++ training data achieves the best performance, also outperforming the similarity-based unsupervised methods. 
This shows that neural models for knowledge selection benefit from supervised training. 
Among different configurations for the RoBERTa classifiers, we can see that training on WOW++  
is the most effective across different metrics. 
Given that the training data size is about the same between WOW++ and WOW-original-subset, the performance gap can be explained by the fact that only a single positive snippet was provided in the latter, whereas multiple positive knowledge sentences are used in WOW++, which is a matched condition for our test setup. 
We also varied the setting for WOW-original-subset by only including one negative example, i.e., total 800 classification training instances. 
This improved the performance slightly, i.e., MRR@1 is 0.35/0.46, but still a large gap with using WOW++ training data. 
Comparing to WOW-original, though WOW++ is much smaller, the model was able to better leverage the information from multiple relevant knowledge snippets and learn the optimal ordering of knowledge candidates.  
The last row in the table again shows that matched training is important. When adding WOW-original to WOW++ for training, the results are not as competitive as just using WOW++ data.

Between the seen and unseen split, the results are generally as expected. However, for the unsupervised methods, we would expect smaller difference since there is no notion of `seen' topics.
One reason for this is that the IDF is in fact learned from the training set, and finetuned RoBERTa retriever also has in-domain unsupervised pre-training. We will investigate the effect of topics further in our future work.

\subsection{Response Generation Results}
For extrinsic evaluation, we perform human evaluation of the responses generated using different knowledge selection methods. We also experimented with computing automatic evaluation metrics such as ROUGE with respect to the human responses in the original WOW dataset but found the results quite low. This is to be expected given the sensitivity of generated responses to the knowledge we provide in our systems.    

First we evaluate the effect of different ground truth, \emph{Wisdom-of-Crowd} and \emph{WoW Original Gold}, on response generation. 
The \emph{Wisdom-of-Crowd} setting uses the most relevant knowledge sentence according to the WOC scores from WOW++, whereas the \emph{WoW Original Gold} setting uses the gold relevant snippet from the original WOW dataset. 
We randomly sampled 100 dialogue contexts for human evaluation. 
Each dialogue context coupled with the system output is provided to an annotator who is asked to evaluate the output according to appropriateness and informativeness.  The responses were evaluated by 2 expert annotators on an ordinal 0-2 scale for each metric. 
Results are provided in Table \ref{tab:human-eval}.
It is clear that the single ground truth in the original WOW data is not as good as the WOC scoring scheme for picking good knowledge snippets to feed to the downstream response generator.

\begin{table}[h]
\centering
\begin{small}
\begin{tabular}{|l|l|l|}
\hline
Knowledge used & \textbf{Appropriate} &   \textbf{Informative} \\ \hline
WOW Original Gold & 1.33/1.60 	 & 1.00/1.18 \\ \hline
Wisdom-of-Crowd & \textbf{1.34}/\textbf{1.70} &   1.25/\textbf{1.29} \\ \hline
\end{tabular}
\caption{Response generation: human evaluation results of responses when ground truth knowledge is provided to the NRG model (Test Unseen/Test seen). }
\label{tab:human-eval}
\end{small}
\end{table}

Next we compare the responses generated using different automatic knowledge selection approaches.  
Based on results from Table~\ref{tab:automatic-eval}, we just ran this human evaluation for two methods, TF-IDF and RoBERTa reranker. 
For this evaluation, since providing absolute scores is more subjective, we performed a preference test by asking annotators to choose which response is better between the two candidates, on two dimensions: appropriate and informative.  
Results in Table~\ref{tab:human-eval-2} show that consistent with the intrinsic knowledge selection results, the RoBERTa model trained on the WOW++ data performs better, showing it is more able to provide relevant knowledge to be used by the downstream response generator. 
One problem we found with the TF-IDF method is that it may select a knowledge sentence that repeats information in the dialog context. This is not surprising since it is heavily relying on lexical overlap, whereas the supervised RoBERTa reranker has learned about both relevance and repetition during training. 
Examples in Appendix show this issue for TF-IDF. 

\begin{table}[h]
\centering
\begin{small}
\begin{tabular}{|l|c|c|}
\hline
Knowledge used & \textbf{Appropriate} & \textbf{Informative} \\ \hline
TF-IDF & 21.5\%   & 22\%  \\ \hline
RoBERTa & 49.5\% & 47.5\% \\ \hline
\end{tabular}
\caption{Response generation using different knowledge selection method: TF-IDF vs. RoBERTa. Results show the percentage of the method chosen as the preferred one for that dimension.}
\label{tab:human-eval-2}
\end{small}
\end{table}



\vspace{-0.25in}
\section{Conclusion}
In this work, 
we demonstrated that knowledge selection is an intrinsically ambiguous task for open-domain conversations, which necessitates improved notions of relevance in our benchmark datasets. We used this insight to propose a new measure for knowledge sentence relevance called \emph{Wisdom-of-the-Crowd} relevance. 
Using this measure, we annotated a new collection of dialogues with relevant knowledge called WOW++. We then evaluated a number of knowledge selection algorithms on our new dataset using both intrinsic and extrinsic metrics, and demonstrate that neural rankers trained leveraging WOW++ can outperform traditional knowledge selection algorithms.  It is worth noting, however, that annotating a knowledge selection dataset with all relevant snippets as we have done for WOW++ is a time-intensive task that may be expensive to scale up. Future work should investigate how to develop more low-resource rankers  or how to bootstrap from a high quality seed dataset like WOW++ to a larger corpus.

\bibliographystyle{acl_natbib}
\bibliography{acl2021}

\appendix

\begin{figure*}
    \centering
 \includegraphics[width=\textwidth]{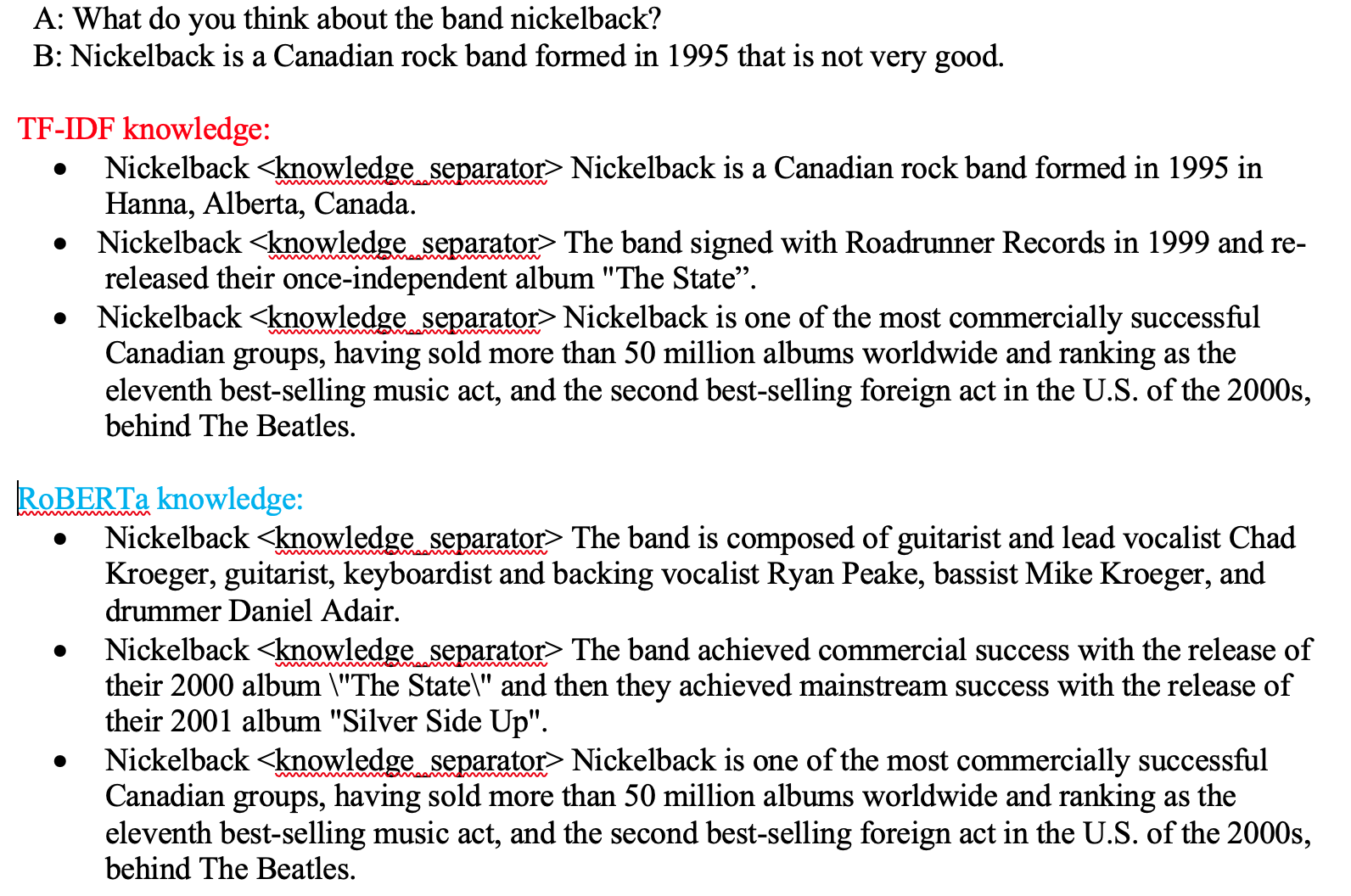}
\caption{Example dialogues showing the top 3 ranked knowledge snippets for both the \emph{Tf-IDF} and \emph{RoBERTa Reranker} models. Note how the \emph{RoBERTa Reranker} tends to select knowledge that is more semantically coherent with the most recent dialogue context. By comparison, the \emph{TF-IDF} model only focuses on snippets with high lexical overlap, resulting in repeated information.}
\end{figure*}

\begin{figure*}
\centering
\includegraphics[width=\textwidth]{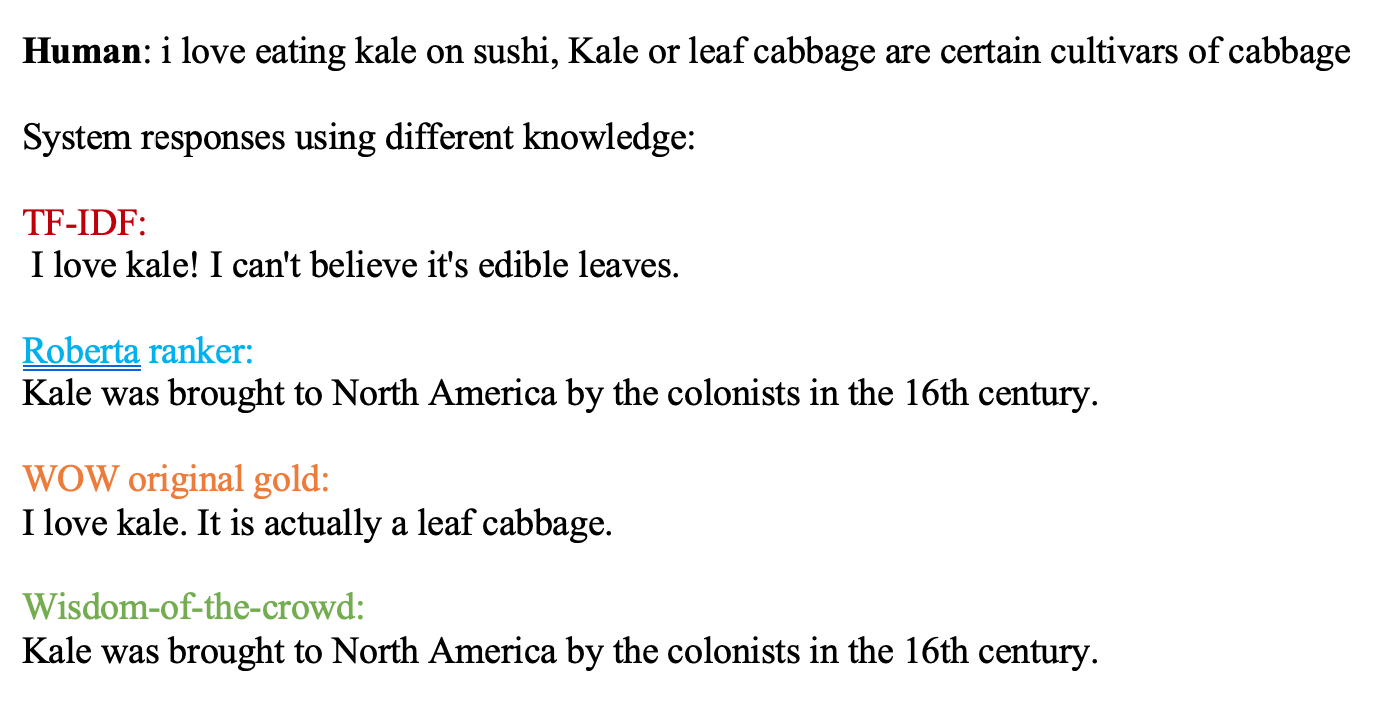}
   \caption{Example dialogue with corresponding responses. Note that both the \emph{TF-IDF} and \emph{WOW original gold} responses repeat information that was previously given by the user's turn.}
\end{figure*}

\begin{figure*}[t]
    \centering
    \includegraphics[width=\textwidth]{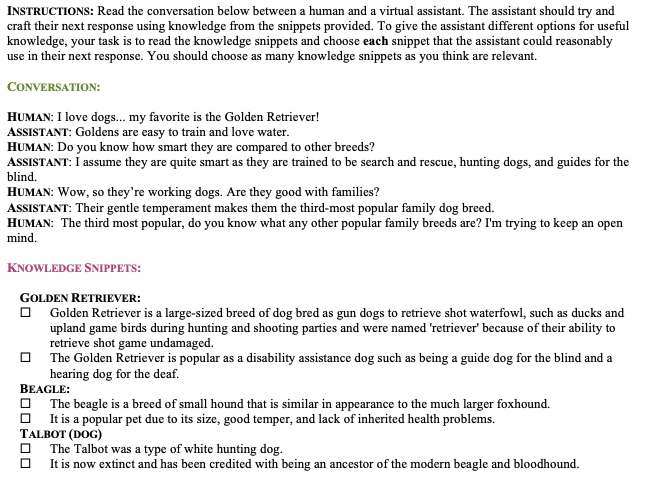}
    \caption{Adapted version of knowledge selection task presented to crowdworkers.}
    \label{fig:ks_instructions}
\end{figure*}

\end{document}